\documentclass[11pt,a4paper]{article}
\usepackage{authblk}
\usepackage[T1]{fontenc}
\usepackage[hyperref]{emnlp-ijcnlp-2019}
\usepackage{times}
\usepackage{comment}
\usepackage{booktabs}
\usepackage{latexsym}
\usepackage{amsmath}
\usepackage{bibentry}
\usepackage{graphicx}
\usepackage{url}
\usepackage{algorithm}
\usepackage{algorithmic}
\usepackage[normalem]{ulem}

\newcommand{\clusterAbf}[1]{\textbf{\color{red}\dotuline{#1}}}
\newcommand{\clusterBbf}[1]{\textbf{\color{blue}\dashuline{#1}}}
\newcommand{\clusterCbf}[1]{\textbf{\color{orange}\uline{#1}}}
\newcommand{\clusterDbf}[1]{\textbf{\color{teal}\uwave{#1}}}
\newcommand{\clusterA}[1]{{\color{red}\dotuline{#1}}}
\newcommand{\clusterB}[1]{{\color{blue}\dashuline{#1}}}
\newcommand{\clusterC}[1]{{\color{orange}\uline{#1}}}

\aclfinalcopy 

\title{Rewarding Coreference Resolvers for Being Consistent \\with World Knowledge}

\author[1]{\textbf{Rahul Aralikatte}}
\author[1]{\textbf{Heather Lent}}
\author[1]{\textbf{Ana Valeria Gonzalez}}
\author[1]{\textbf{Daniel Hershcovich}}
\author[1,2]{\\\textbf{Chen Qiu}}
\author[3]{\textbf{Anders Sandholm}}
\author[3]{\textbf{Michael Ringaard}}
\author[1,3]{\textbf{Anders S{\o}gaard}}
\affil[ ]{$^1$University of Copenhagen, $^2$China University of Geosciences, $^3$Google Research}
\affil[ ]{\texttt{\{rahul,hcl,ana,hershcovich,chen.qiu,soegaard\}@di.ku.dk},}
\affil[ ]{\texttt{\{sandholm,ringgaard\}@google.com}}

\begin{document}
\maketitle
\begin{abstract}
Unresolved coreference is a bottleneck for relation extraction, and high-quality coreference resolvers may  produce an output that makes it a lot easier to extract knowledge triples. We show how to improve coreference resolvers by forwarding their input to a relation extraction system and reward the resolvers for producing triples that are found in knowledge bases. Since relation extraction systems can rely on different forms of supervision and be biased in different ways, we obtain the best performance, improving over the state of the art, using multi-task reinforcement learning.
\end{abstract}

\section{Introduction}\label{sec:introduction}

Coreference annotations are costly and difficult to obtain, since trained annotators with sufficient world knowledge are necessary for reliable annotations. This paper presents a way to {\em simulate} annotators using reinforcement learning. To motivate our approach, we rely on the following example from \newcite[colors added to mark entity mentions]{strube-etal}:
\begin{itemize}
    \item[(1)] [\clusterAbf{Lynyrd Skynyrd}]$_1$ was formed in \clusterBbf{Florida$_2$}. Other bands from [\clusterBbf{the Sunshine State}]$_2$ include \clusterCbf{Fireflight} and \clusterDbf{Marilyn Manson}.
\end{itemize}

\newcite{strube-etal} cite the association between \clusterB{Florida} and \clusterB{the Sunshine State} as an example of a common source of name-name recall error for state-of-the-art coreference resolution systems. The challenge is that the two names co-occur relatively infrequently and are unlikely to do so in a moderate-sized, manually annotated training corpus. A state-of-the-art system may be able to infer the relation using distributional information about the phrase \clusterB{the Sunshine State}, but is likely to have limited evidence for the decision that it is coreferential with \clusterB{Florida} rather than \clusterA{Lynyrd Skynyrd}.

While coreference-annotated data is scarce, knowledge bases including factual information (such as that \clusterC{Fireflight} is from \clusterB{Florida}) are increasingly available. For a human annotator unaware that \clusterB{Florida} is sometimes referred to as \clusterB{the Sunshine State}, the information that \clusterC{Fireflight} is from \clusterB{Florida} is sufficient to establish that \clusterB{Florida} and \clusterB{the Sunshine State} are (with high probability) coreferential. This paper explores a novel architecture for making use of such information from knowledge bases by tying a coreference resolution system to a relation extraction system, enabling us to reward the coreference system for making predictions that lead us to infer facts that are consistent with such knowledge bases. This potentially provides us with more evidence for resolving coreference such as (1). 

\begin{figure}
    \centering
    \includegraphics[width=\columnwidth]{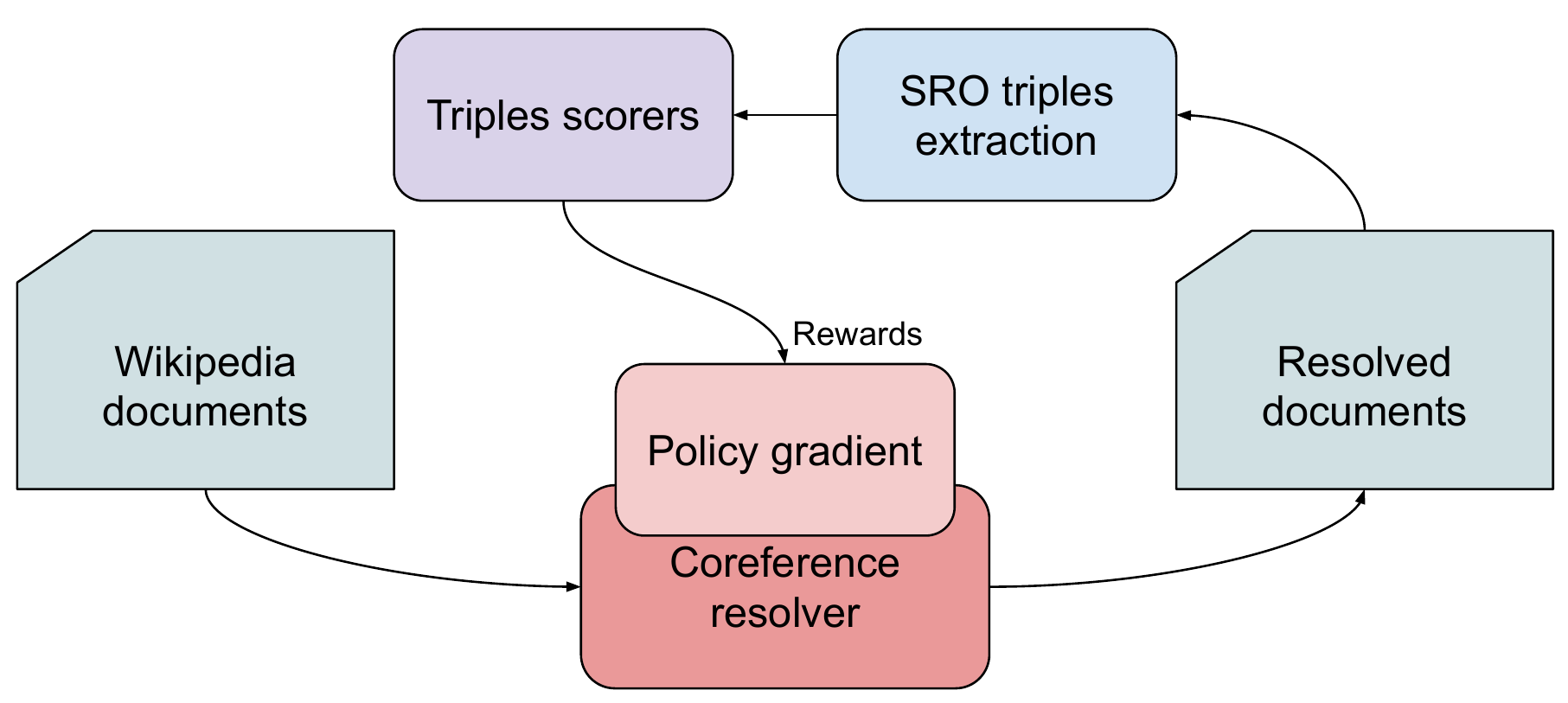}
    \caption{\label{fig:sys_arch}Our strategy for training a coreference resolver using reward from relation extraction.}
\end{figure}

\begin{figure*}
    \centering
    \includegraphics[width=\textwidth]{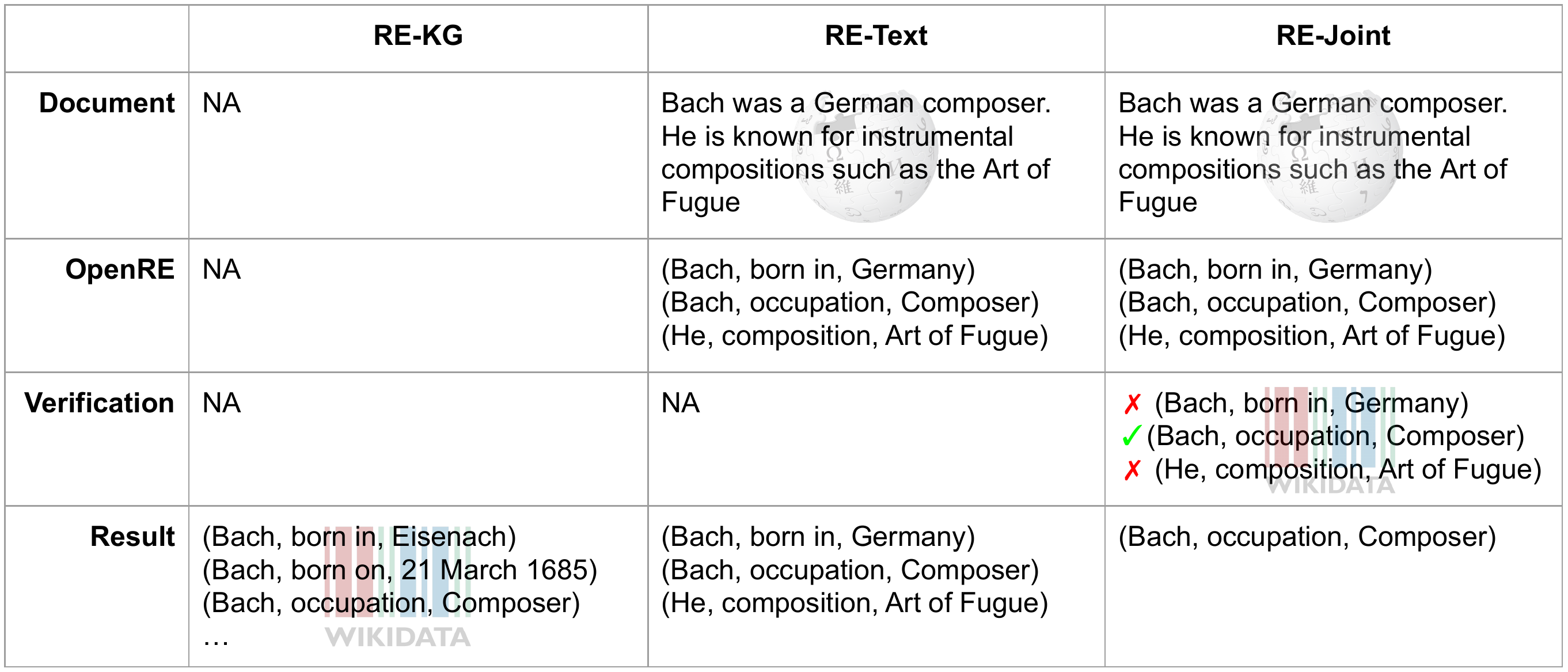}
    \caption{\label{fig:overall}The columns show the different pipelines used to obtain data for training the reward models. The pipeline for: (i) RE-KG directly extracts triples from Wikidata, (ii) RE-Text runs Wikipedia summaries through OpenRE to generate triples, and (iii) RE-Joint adds an additional verification step by checking if the generated triples exist in Wikidata.}
\end{figure*}


We propose a training strategy (Figure~\ref{fig:sys_arch}) in which we pass on the predictions of a neural coreference resolver to an open relation extraction (OpenRE) system, matching relations extracted from resolved sentences with a knowledge base. We show how checking the produced relationships for consistency against the knowledge base produces a reward that is, indirectly, a signal about the quality of the coreference resolution. In order to generalize this signal beyond the coverage of the knowledge base, we train a Universal Schema model \cite{riedel-etal-2013-relation} and use its confidence as our reward function.  With this reward function, we do policy-gradient fine-tuning of our coreference resolver, effectively optimizing its predictions' consistency with world knowledge. 

\paragraph{Contributions}
We demonstrate that training a coreference resolver by reinforcement learning with rewards from a relation extraction system, results in improvements for coreference resolution.
Our code is made publicly available at \url{https://github.com/rahular/coref-rl}

\section{Consistency Reward for Coreference Resolution}\label{sec:strategy}

In order to reward a coreference resolver for being consistent with world knowledge, we propose a simple training strategy based on relation extraction: (i) Sample a Wikipedia\footnote{\url{https://www.wikipedia.org}} document at random, (ii)  Replace mentions with their antecedents using a coreference resolver, (iii)  Apply an off-the-shelf openRE system to each rewritten document, (iv) Score relationships that include coreferent mentions using Universal Schema, and (v)  Use the score as a reward for training the coreference resolvers. 

\paragraph{Reward functions} To model consistency with world knowledge, we train different Universal Schema models \cite{riedel-etal-2013-relation,Verga:McCallum:16}, resulting in three reward functions (Figure~\ref{fig:overall}): \textbf{RE-KG} (Knowledge Graph Universal Schema) is trained to predict whether two entities are linked in Wikidata\footnote{\url{https://www.wikidata.org}}; \textbf{RE-Text} (Text-based Universal Schema) is trained to predict whether two entities co-occur in Wikipedia; and \textbf{RE-Joint} (Joint Universal Schema) is trained to predict whether two entities are linked {\it and} co-occur. The three rewards focus on different aspects of relationships between entities, giving complimentary views of what entities are related.

Similar to \newcite{verga-etal-2016-multilingual}, we parameterize candidate relation phrases with a BiLSTM \cite{graves2005framewise}, and use pre-trained Wikidata BigGraph embeddings \cite{pbg} as the entity representations. We apply a one-layer MLP on the concatenated representations to get the reward value.

\paragraph{Updating the coreference resolver} Each resolved document is converted into $n$ subject-relation-object (SRO) triples by an open information retrieval system \cite{openIE}. Each triple $t_i$ is then scored using a reward function to obtain a reward $r_i$ for $i \in \{1, \ldots, n\}$. The final document-level reward is the normalized sum of the individual rewards as shown in Equation~\ref{eqn:reward}, where $R_h$ is a moving window containing the previous $h=100$ normalized reward values.

\begin{equation}
    R = \frac{\sum_{i} r_i - mean(R_h)}{stddev(R_h)}
    \label{eqn:reward}
\end{equation}

Since $R$ is not differentiable with respect to the coreference resolver's parameters, we use policy gradient training to update the coreference resolver. We select the best action according to the current policy, using random exploration of the alternative solutions with $p=\frac{1}{10}$. 

\paragraph{Multi-task reinforcement learning} Our overall  training procedure is presented in Algorithm~\ref{alg:distral}. After training the three aforementioned reward models, we create \textbf{RE-Distill} by interpolating their trained weights. Next, we pre-train a coreference resolver using supervised learning, and fine-tune it using each of the three reward functions to get three different coreference policies: \textbf{Coref-KG}, \textbf{Coref-Text} and \textbf{Coref-Joint}, respectively. We then use multi-task reinforcement learning to combine these three policies to get \textbf{Coref-Distill}. Our approach is a particular instance of DisTraL \cite{distral}, using policy gradient and model interpolation.  Finally, \textbf{Coref-Distill} is fine-tuned with rewards from {\bf RE-Distill}. 

\begin{algorithm}
\begin{small}
\caption{\label{alg:distral} Multi-task Reinforcement Learning}
\begin{algorithmic}
\REQUIRE  Baseline initialized policies $\theta_n$ for $n\in \{1,2,3\}$ \label{alg:pretrained-policies}\\
\REQUIRE Reward functions \texttt{reward$_n$} for $n\in \{1,2,3\}$
\REQUIRE Distilled reward function \texttt{reward$_*$}
\WHILE{stopping criterion not met}
\STATE Sample $k$ documents $D^k$
\FOR{$d\in D^k$}
\FOR{$n\in\{1,2,3\}$}
\STATE $\mathcal{C}_d$ = entity clusters with $\theta_n$
\STATE $d'$ = resolve $d$ with $\mathcal{C}_d$
\STATE $\mathcal{T}$ = obtain OpenIE triples for $d'$ 
\STATE $r$ = reward$_n$($d'$)
\STATE $\hat{g}_k$ = policy gradient for $\theta_n$ with reward $r$
\STATE 
$\theta_n^{k+1}=\theta_n^k+\alpha_k\hat{g}_k$
\ENDFOR
\ENDFOR
\ENDWHILE 
\STATE Distilled policy $\theta_*=\frac{\theta_1+\theta_2+\theta_3}{3}$
\STATE Sample $k$ documents $D^k$
\FOR{$d\in D^k$}
\STATE $d'$ = resolve $d$ with $\mathcal{C}_d$
\STATE $\mathcal{T}$ = obtain OpenIE triples for $d'$ 
\STATE $r$ = reward$_*$($d'$)
\STATE $\hat{g}_k$ = policy gradient for $\theta_*$ with reward $r$
\STATE 
$\theta_*^{k+1}=\theta_*^k+\alpha_k\hat{g}_k$
\ENDFOR
\RETURN Distilled policy $\theta_*$
\end{algorithmic}
\end{small}
\end{algorithm}

\section{Experiments}\label{sec:experiments}

We use a state-of-the-art neural coreference resolution model \cite{lee2018higher} as our baseline coreference resolver.\footnote{\url{https://github.com/kentonl/e2e-coref}} This model extends \citet{lee2017end} with coarse-to-fine inference and ELMo pretrained embeddings \cite{peters2018deep}.


\paragraph{Data} We use the standard training, validation, and test splits from the English OntoNotes.\footnote{\url{https://catalog.ldc.upenn.edu/LDC2013T19}}
We also evaluate on the English WikiCoref \cite{wikicoref}, with a validation and test split of 10 and 20 documents respectively.

\paragraph{Reward model training} We use data from English Wikipedia and Wikidata to train our three reward models.
For training \textbf{RE-KG}, we sample 1 million Wikidata triples, and expand them to 12 million triples by replacing relation phrases with their aliases.
For \textbf{RE-Text}, we pass the summary paragraphs from 50,000 random Wikipedia pages to Stanford's OpenIE extractor \cite{corenlp}, creating 2 million triples.
For \textbf{RE-Joint}, we only use Wikipedia triples that are grounded in Wikidata, resulting in 60,000 triples.\footnote{That is, we retain only those triples whose subject and object can be linked to an entity in Wikidata.}
We further sample 200,000 triples from Wikidata and Wikipedia for validation, and train the reward models with early stopping based on the F$_1$ score of their predictions.

\paragraph{Evaluation}
All models are evaluated using the standard CoNLL metric, which is the average F$_1$ score over MUC, CEAFe, and $B^3$ \cite{denis2009global}.

\section{Results}\label{sec:results}

Since the quality of our reward models is essential to the performance of the coreference resolver adaptations, we first report the validation accuracy and F$_1$ scores of the four reward models used, in Table~\ref{tab:reward_results}. We clearly see the advantage of distillation, with a 5\% absolute difference between the best single model ({\bf RE-Text}) and {\bf RE-Distill}.

\begin{table}[t]
    \centering
    \begin{tabular}{l|ccc}
        \toprule
        {\bf System} & {\bf Data} & {\bf Accuracy} & {\bf F$_1$ score} \\
        \midrule
        RE-KG & 12M & 0.64 & 0.78 \\
        RE-Text & 2M & 0.71 & 0.83 \\
        RE-Joint & 60K & 0.58 & 0.73 \\
        \midrule
        RE-Distill & --- & \textbf{0.78} & \textbf{0.88} \\
        \bottomrule
    \end{tabular}
    \caption{Training data size, accuracy and F$_1$ scores of the reward models on the 200,000 validation triples.\label{tab:reward_results}}
\end{table}

\begin{table}[t]
    \centering
    \begin{tabular}{l|cc}
        \toprule
        {\bf System} & {\bf OntoNotes}&{\bf WikiCoref} \\
        \midrule
        \newcite{lee2018higher} & 72.60 & 57.49 \\
        \midrule
        Coref-KG & 72.96 & 57.84 \\
        Coref-Text & 72.99 & 57.54 \\
        Coref-Joint & 72.77 & 57.51 \\
        \midrule
        Coref-Distill & \textbf{73.10} & \textbf{58.14} \\
        \bottomrule
    \end{tabular}
    \caption{Coreference results: average F$_1$ scores on the OntoNotes and WikiCoref test sets. Differences are significant w.r.t. $B^3$ (bootstrap test, $p<0.05$).\label{tab:coref_results}} 
\end{table}

Table~\ref{tab:coref_results} presents the downstream effects of applying these reward functions to our baseline coreference policy.\footnote{The models were re-trained from scratch, and the scores are slightly different from those reported in \newcite{lee2018higher}.}

The coreference resolution results are similar to the relation extraction results: using a distilled policy, learned through multi-task reinforcement learning, leads to better results on both datasets.\footnote{We repeat this experiment three times with different random seeds and observed the same pattern and very robust performance across the board.}

While improvements over the current state of the art are relatively small, they reflect significant progress, as they demonstrate the ability to successfully augment coreference resolvers with ``free" data from large-scale KB like Wikidata. For relation extraction, this could have positive downstream effects, and also ensure that relations are consistent with real world knowledge. Moreover, this approach has the potential to also be beneficial for coreference resolution in low resource languages, where less annotated data is available, as Wikidata triples are abundant for many languages.

\section{Analysis}\label{sec:analysis}
\begin{figure*}
    \centering
    \includegraphics[width=\textwidth]{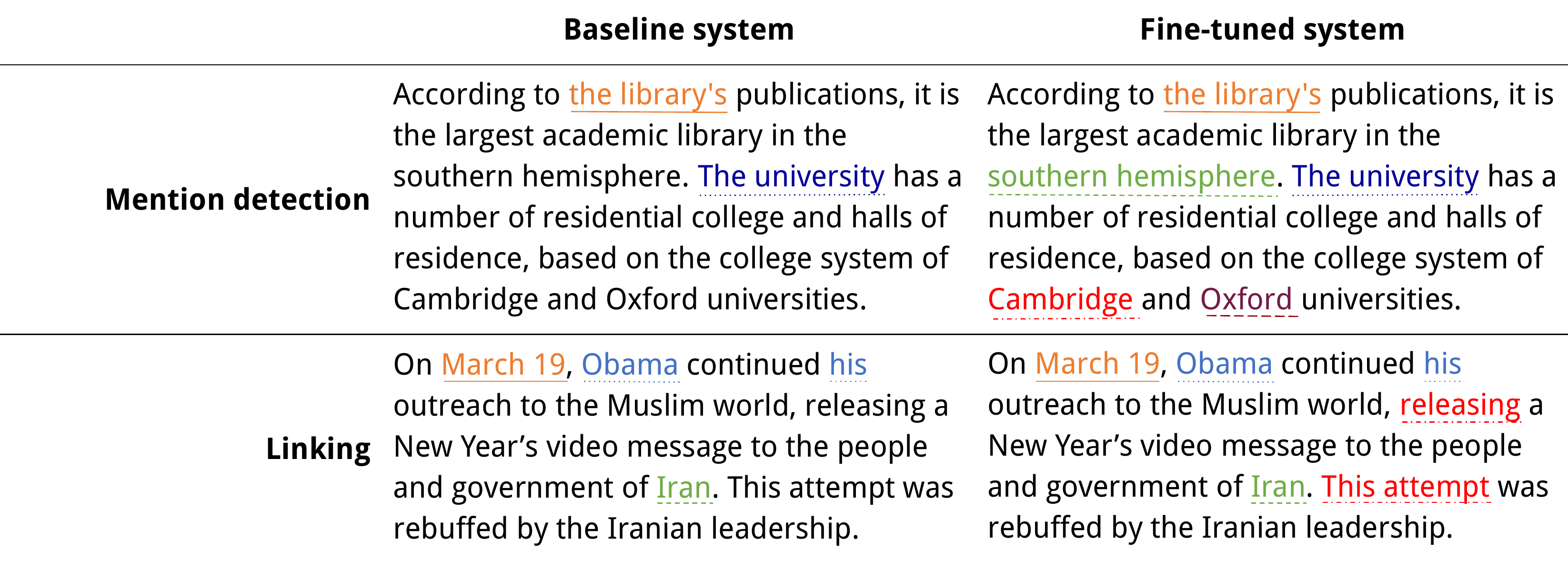}
    \caption{\label{fig:analysis}Mention detection and linking examples by the baseline system from \newcite{lee2018higher}, and the best performing fine-tuned system (Coref-Distill). Mentions of the same color are linked to form a coreference cluster.}
\end{figure*}

Empirically, we find that fine-tuning the coreference resolver on Wikidata results in two kinds of improvements: 

\paragraph{Better mention detection} Since the model is rewarded if the SRO triples produced from the resolved document are present in Wikidata, the model can do well only if it correctly resolves the subject and object, which are usually named entities (more generally, noun phrases). Indeed, we see an improvement in mention detection as exemplified in the first example of Figure~\ref{fig:analysis}. Compared to the baseline, the fine-tuned model identifies a larger number of entities, including ``southern hemisphere'', ``Cambridge'' and ``Oxford'', which are missed by the baseline model.

\paragraph{Better linking}  As a direct consequence of the above, the model is inclined to also link noun phrases that are not entities. In the second example of Figure~\ref{fig:analysis}, we see that ``This attempt'' is linked to ``releasing'' by the fine-tuned model. Interestingly, we do not see this type of \textit{eventive} noun phrase linking either in OntoNotes or in the predictions of the baseline model. 

This phenomenon, however, also has a side-effect of producing singleton clusters and spurious linking, which adversely affect the recall. On the OntoNotes test data, while the average precision of the best performing fine-tuned model is higher than the baseline (75.62 vs. 73.80), a drop in recall (70.75 vs. 71.34) causes the final F$_1$ score to only marginally improve.

\section{Related Work}\label{sec:related}
\paragraph{Coreference resolution} Among neural coreference resolvers \cite{Wu2017ADL, Meng2018TriadbasedNN}, \citet{lee2017end} were the first to propose an end-to-end resolver which did not rely on hand-crafted rules or a syntactic parser. Extending this work, \citet{lee2018higher} introduced a novel attention mechanism for iteratively ranking spans of candidate coreferent mentions, thereby improving the identification of long distance coreference chains. \newcite{zhang-etal} improve pronoun coreference resolution by 2.2 F1 using linguistic features (gender, animacy and plurality) and a frequency based predicate-argument selection preference as external knowledge. \newcite{emami-etal} incorporate knowledge into coreference resolution by means of information retrieval, finding sentences that are syntactically similar to a given instance, and improving F1 by 0.16.

\paragraph{Reinforcement learning} RL has been used for many NLP tasks, including coreference resolution \cite{clark2016deep} and relation extraction \cite{Zeng2018LargeSR}. \citet{clark2016deep} use RL to improve coreference resolution by optimizing their mention ranking model and directly use the standard evaluation metrics as the rewards. We, on the other hand, perform end-to-end optimization by rewarding the model's consistency with real world knowledge using relation extraction. To our knowledge, we are the first to use consistency with world knowledge as a reward for tasks other than knowledge base construction.\footnote{\newcite{Mao:ea:18}, for example, use reinforcement learning with consistency-like reward to induce lexical taxonomies.}  

\paragraph{Knowledge bases} Knowledge bases have been leveraged across multiple tasks across NLP \cite{Bordes2011LearningSE,Chang2014TypedTD, Lin2015ModelingRP, Toutanova2015RepresentingTF, Yang2017LeveragingKB}. Specifically for coreference resolution, \citet{Prokofyev2015SANAP} implement a resolver that ensures semantic relatedness of resulting coreference clusters by leveraging Semantic Web annotations. Their work incorporates knowledge graph information only in the final stage of the resolver's pipeline, and not during training. In contrast, our work augments information from the knowledge base directly into the training pipeline. Also, they use DBpedia \cite{dbpedia07} as the ontology. Although both Wikidata and DBpedia are designed to support working with Wikipedia articles, DBpedia can be considered as a subset of Wikidata as Wikipedia infoboxes are its main data source. The advantage of Wikidata over DBpedia is its size, and the fact that it is multilingual, which will allow applying our method to other languages in the future. 

\section{Conclusion}\label{sec:conclusion}

We presented an architecture for adapting coreference resolvers by rewarding them for being consistent with world knowledge. Using simple multi-task reinforcement learning and a knowledge extraction pipeline, we achieved  improvements over the state of the art across two datasets. We believe this is an important first step in exploring the usefulness of knowledge bases in the context of coreference resolution and other discourse-level phenomena. In this area, manually annotated data is particularly expensive, and we believe leveraging knowledge bases will eventually reduce the need for manual annotation. 

\section*{Acknowlegments}

We thank the reviewers for their valuable comments.
Rahul Aralikatte, Daniel Hershcovich, Heather Lent, and Anders S{\o}gaard are funded by a Google Focused Research Award. Heather Lent is also funded by the European Union's Horizon 2020 research and innovation programme under the Marie Sk{\l}odowska-Curie grant agreement No. 801199. Chen Qiu is funded in part by the National Natural Science Foundation of China under grant No. 61773355 and the China Scholarship Council.

\bibliography{coref-rl}

\begin{thebibliography}{28}
\expandafter\ifx\csname natexlab\endcsname\relax\def\natexlab#1{#1}\fi

\bibitem[{Angeli et~al.(2015)Angeli, Johnson~Premkumar, and Manning}]{openIE}
Gabor Angeli, Melvin~Jose Johnson~Premkumar, and Christopher~D. Manning. 2015.
\newblock \href {https://doi.org/10.3115/v1/P15-1034} {Leveraging linguistic
  structure for open domain information extraction}.
\newblock In \emph{Proceedings of the 53rd Annual Meeting of the Association
  for Computational Linguistics and the 7th International Joint Conference on
  Natural Language Processing (Volume 1: Long Papers)}, pages 344--354,
  Beijing, China. Association for Computational Linguistics.

\bibitem[{Auer et~al.(2007)Auer, Bizer, Kobilarov, Lehmann, Cyganiak, and
  Ives}]{dbpedia07}
S\"{o}ren Auer, Christian Bizer, Georgi Kobilarov, Jens Lehmann, Richard
  Cyganiak, and Zachary Ives. 2007.
\newblock \href {http://dl.acm.org/citation.cfm?id=1785162.1785216} {Dbpedia: A
  nucleus for a web of open data}.
\newblock In \emph{Proceedings of the 6th International The Semantic Web and
  2Nd Asian Conference on Asian Semantic Web Conference}, ISWC'07/ASWC'07,
  pages 722--735, Berlin, Heidelberg. Springer-Verlag.

\bibitem[{Bordes et~al.(2011)Bordes, Weston, Collobert, and
  Bengio}]{Bordes2011LearningSE}
Antoine Bordes, Jason Weston, Ronan Collobert, and Yoshua Bengio. 2011.
\newblock Learning structured embeddings of knowledge bases.
\newblock In \emph{AAAI}.

\bibitem[{Chang et~al.(2014)Chang, tau Yih, Yang, and Meek}]{Chang2014TypedTD}
Kai-Wei Chang, Wen tau Yih, Bishan Yang, and Christopher Meek. 2014.
\newblock Typed tensor decomposition of knowledge bases for relation
  extraction.
\newblock In \emph{EMNLP}.

\bibitem[{Clark and Manning(2016)}]{clark2016deep}
Kevin Clark and Christopher~D Manning. 2016.
\newblock Deep reinforcement learning for mention-ranking coreference models.
\newblock In \emph{Proceedings of the 2016 Conference on Empirical Methods in
  Natural Language Processing}, pages 2256--2262.

\bibitem[{Denis and Baldridge(2009)}]{denis2009global}
Pascal Denis and Jason Baldridge. 2009.
\newblock Global joint models for coreference resolution and named entity
  classification.
\newblock \emph{Procesamiento del Lenguaje Natural}, 42.

\bibitem[{Emami et~al.(2018)Emami, Trischler, Suleman, and Cheung}]{emami-etal}
Ali Emami, Adam Trischler, Kaheer Suleman, and Jackie Chi~Kit Cheung. 2018.
\newblock \href {https://doi.org/10.18653/v1/N18-4004} {A generalized knowledge
  hunting framework for the {W}inograd schema challenge}.
\newblock In \emph{Proceedings of the 2018 Conference of the North {A}merican
  Chapter of the Association for Computational Linguistics: Student Research
  Workshop}, pages 25--31, New Orleans, Louisiana, USA. Association for
  Computational Linguistics.

\bibitem[{Ghaddar and Langlais(2016)}]{wikicoref}
Abbas Ghaddar and Philippe Langlais. 2016.
\newblock Wikicoref: An english coreference-annotated corpus of wikipedia
  articles.
\newblock In \emph{Proceedings of the Tenth International Conference on
  Language Resources and Evaluation (LREC 2016)}, Portoro{\v z}, Slovenia.
  European Language Resources Association (ELRA), European Language Resources
  Association (ELRA).

\bibitem[{Graves and Schmidhuber(2005)}]{graves2005framewise}
Alex Graves and J{\"u}rgen Schmidhuber. 2005.
\newblock Framewise phoneme classification with bidirectional lstm and other
  neural network architectures.
\newblock \emph{Neural Networks}, 18(5-6):602--610.

\bibitem[{Lee et~al.(2017)Lee, He, Lewis, and Zettlemoyer}]{lee2017end}
Kenton Lee, Luheng He, Mike Lewis, and Luke Zettlemoyer. 2017.
\newblock End-to-end neural coreference resolution.
\newblock In \emph{Proceedings of the 2017 Conference on Empirical Methods in
  Natural Language Processing}, pages 188--197.

\bibitem[{Lee et~al.(2018)Lee, He, and Zettlemoyer}]{lee2018higher}
Kenton Lee, Luheng He, and Luke Zettlemoyer. 2018.
\newblock Higher-order coreference resolution with coarse-to-fine inference.
\newblock In \emph{Proceedings of the 2018 Conference of the North American
  Chapter of the Association for Computational Linguistics: Human Language
  Technologies, Volume 2 (Short Papers)}, pages 687--692.

\bibitem[{Lerer et~al.(2019)Lerer, Wu, Shen, Lacroix, Wehrstedt, Bose, and
  Peysakhovich}]{pbg}
Adam Lerer, Ledell Wu, Jiajun Shen, Timothee Lacroix, Luca Wehrstedt, Abhijit
  Bose, and Alex Peysakhovich. 2019.
\newblock {PyTorch-BigGraph: A Large-scale Graph Embedding System}.
\newblock In \emph{Proceedings of the 2nd SysML Conference}, Palo Alto, CA,
  USA.

\bibitem[{Lin et~al.(2015)Lin, Liu, and Sun}]{Lin2015ModelingRP}
Yankai Lin, Zhiyuan Liu, and Maosong Sun. 2015.
\newblock Modeling relation paths for representation learning of knowledge
  bases.
\newblock In \emph{EMNLP}.

\bibitem[{Manning et~al.(2014)Manning, Surdeanu, Bauer, Finkel, Bethard, and
  McClosky}]{corenlp}
Christopher~D. Manning, Mihai Surdeanu, John Bauer, Jenny Finkel, Steven~J.
  Bethard, and David McClosky. 2014.
\newblock \href {http://www.aclweb.org/anthology/P/P14/P14-5010} {The
  {Stanford} {CoreNLP} natural language processing toolkit}.
\newblock In \emph{Association for Computational Linguistics (ACL) System
  Demonstrations}, pages 55--60.

\bibitem[{Mao et~al.(2018)Mao, Ren, Shen, Gu, and Han}]{Mao:ea:18}
Yuning Mao, Xiang Ren, Jiaming Shen, Xiaotao Gu, and Jiawei Han. 2018.
\newblock Building a large-scale annotated chinese corpus.
\newblock In \emph{ACL}.

\bibitem[{Martschat and Strube(2014)}]{strube-etal}
Sebastian Martschat and Michael Strube. 2014.
\newblock \href {https://doi.org/10.3115/v1/D14-1221} {Recall error analysis
  for coreference resolution}.
\newblock In \emph{Proceedings of the 2014 Conference on Empirical Methods in
  Natural Language Processing ({EMNLP})}, pages 2070--2081, Doha, Qatar.
  Association for Computational Linguistics.

\bibitem[{Meng and Rumshisky(2018)}]{Meng2018TriadbasedNN}
Yuanliang Meng and Anna Rumshisky. 2018.
\newblock Triad-based neural network for coreference resolution.
\newblock In \emph{COLING}.

\bibitem[{Peters et~al.(2018)Peters, Neumann, Iyyer, Gardner, Clark, Lee, and
  Zettlemoyer}]{peters2018deep}
Matthew Peters, Mark Neumann, Mohit Iyyer, Matt Gardner, Christopher Clark,
  Kenton Lee, and Luke Zettlemoyer. 2018.
\newblock Deep contextualized word representations.
\newblock In \emph{Proceedings of the 2018 Conference of the North American
  Chapter of the Association for Computational Linguistics: Human Language
  Technologies, Volume 1 (Long Papers)}, pages 2227--2237.

\bibitem[{Prokofyev et~al.(2015)Prokofyev, Tonon, Luggen, Vouilloz, Difallah,
  and Cudr{\'e}-Mauroux}]{Prokofyev2015SANAP}
Roman Prokofyev, Alberto Tonon, Michael Luggen, Loic Vouilloz, Djellel~Eddine
  Difallah, and Philippe Cudr{\'e}-Mauroux. 2015.
\newblock Sanaphor: Ontology-based coreference resolution.
\newblock In \emph{International Semantic Web Conference}.

\bibitem[{Riedel et~al.(2013)Riedel, Yao, McCallum, and
  Marlin}]{riedel-etal-2013-relation}
Sebastian Riedel, Limin Yao, Andrew McCallum, and Benjamin~M. Marlin. 2013.
\newblock \href {https://www.aclweb.org/anthology/N13-1008} {Relation
  extraction with matrix factorization and universal schemas}.
\newblock In \emph{Proceedings of the 2013 Conference of the North {A}merican
  Chapter of the Association for Computational Linguistics: Human Language
  Technologies}, pages 74--84, Atlanta, Georgia. Association for Computational
  Linguistics.

\bibitem[{Teh et~al.(2017)Teh, Bapst, Czarnecki, Quan, Kirkpatrick, Hadsell,
  Heess, and Pascanu}]{distral}
Yee Teh, Victor Bapst, Wojciech~M. Czarnecki, John Quan, James Kirkpatrick,
  Raia Hadsell, Nicolas Heess, and Razvan Pascanu. 2017.
\newblock \href
  {http://papers.nips.cc/paper/7036-distral-robust-multitask-reinforcement-learning.pdf}
  {Distral: Robust multitask reinforcement learning}.
\newblock In I.~Guyon, U.~V. Luxburg, S.~Bengio, H.~Wallach, R.~Fergus,
  S.~Vishwanathan, and R.~Garnett, editors, \emph{Advances in Neural
  Information Processing Systems 30}, pages 4496--4506. Curran Associates, Inc.

\bibitem[{Toutanova et~al.(2015)Toutanova, Chen, Pantel, Poon, Choudhury, and
  Gamon}]{Toutanova2015RepresentingTF}
Kristina Toutanova, Danqi Chen, Patrick Pantel, Hoifung Poon, Pallavi
  Choudhury, and Michael Gamon. 2015.
\newblock Representing text for joint embedding of text and knowledge bases.
\newblock In \emph{EMNLP}.

\bibitem[{Verga et~al.(2016)Verga, Belanger, Strubell, Roth, and
  McCallum}]{verga-etal-2016-multilingual}
Patrick Verga, David Belanger, Emma Strubell, Benjamin Roth, and Andrew
  McCallum. 2016.
\newblock \href {https://doi.org/10.18653/v1/N16-1103} {Multilingual relation
  extraction using compositional universal schema}.
\newblock In \emph{Proceedings of the 2016 Conference of the North {A}merican
  Chapter of the Association for Computational Linguistics: Human Language
  Technologies}, pages 886--896, San Diego, California. Association for
  Computational Linguistics.

\bibitem[{Verga and McCallum(2016)}]{Verga:McCallum:16}
Patrick Verga and Andrew McCallum. 2016.
\newblock Row-less universal schema.
\newblock In \emph{AKBC}.

\bibitem[{Wu and Ma(2017)}]{Wu2017ADL}
Jheng-Long Wu and Wei-Yun Ma. 2017.
\newblock A deep learning framework for coreference resolution based on
  convolutional neural network.
\newblock \emph{2017 IEEE 11th International Conference on Semantic Computing
  (ICSC)}, pages 61--64.

\bibitem[{Yang and Mitchell(2017)}]{Yang2017LeveragingKB}
Bishan Yang and Tom~Michael Mitchell. 2017.
\newblock Leveraging knowledge bases in lstms for improving machine reading.
\newblock In \emph{ACL}.

\bibitem[{Zeng et~al.(2018)Zeng, He, Liu, and Zhao}]{Zeng2018LargeSR}
Xiangrong Zeng, Shizhu He, Kang Liu, and Jian Zhao. 2018.
\newblock Large scaled relation extraction with reinforcement learning.
\newblock In \emph{AAAI}.

\bibitem[{Zhang et~al.(2019)Zhang, Song, and Song}]{zhang-etal}
Hongming Zhang, Yan Song, and Yangqiu Song. 2019.
\newblock \href {https://doi.org/10.18653/v1/N19-1093} {Incorporating context
  and external knowledge for pronoun coreference resolution}.
\newblock In \emph{Proceedings of the 2019 Conference of the North {A}merican
  Chapter of the Association for Computational Linguistics: Human Language
  Technologies, Volume 1 (Long and Short Papers)}, pages 872--881, Minneapolis,
  Minnesota. Association for Computational Linguistics.

\end{thebibliography}
\bibliographystyle{acl_natbib}

\end{document}